\documentclass{article}

\usepackage{PRIMEarxiv}

\usepackage{array}
\usepackage{multirow}
\usepackage{pifont}

\usepackage[utf8]{inputenc} 
\usepackage[T1]{fontenc}    
\usepackage{hyperref}       
\usepackage{url}            
\usepackage{booktabs}       
\usepackage{amsfonts}       
\usepackage{nicefrac}       
\usepackage{microtype}      
\usepackage{fancyhdr}       
\usepackage{graphicx}       

\graphicspath{{media/}}     

\newcommand{\orcidicon}{\includegraphics[width=0.32cm]{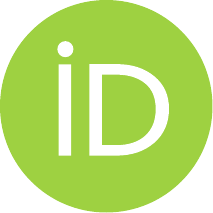}}

\title{Application-oriented automatic hyperparameter optimization for spiking neural network prototyping}

\newcommand{\orcid}[1]{\orcidicon \hspace{0.1em} {#1}}

\author{
  Vittorio Fra\\
  \orcid{0000-0001-9175-2838}\\
  Politecnico di Torino\\
  Turin, Italy\\
  \texttt{vittorio.fra@polito.it}\\
}

\usepackage{glossaries}

\newacronym{ai}{AI}{artificial intelligence}
\newacronym{ann}{ANN}{artificial neural network}
\newacronym{dl}{DL}{deep learning}
\newacronym{hpo}{HPO}{hyperparameter optimization}
\newacronym{ml}{ML}{machine learning}
\newacronym{nni}{NNI}{Neural Network Intelligence}
\newacronym{rsnn}{RSNN}{recurrent SNN}
\newacronym{snn}{SNN}{spiking neural network}
\usepackage[listings,many]{tcolorbox}
\usetikzlibrary{calc}
\tcbuselibrary{breakable}

\definecolor{bluPoli}{RGB}{0,43,73}

\tcbset{listing engine=listings}
\tcbset{mystyle/.style={
  breakable,
  enhanced,
  leftrule=1pt,
  rightrule=1pt,
  toprule=1pt,
  bottomrule=1pt,
  outer arc=0pt,
  arc=0pt,
  colframe=bluPoli,
  colback=white,
  attach boxed title to top left,
  boxed title style={
    colback=bluPoli,
    outer arc=0pt,
    arc=0pt,
    top=1pt,
    bottom=1pt,
    },
  },
}

\newtcblisting[auto counter]{mycodebox}[2][]{
    mystyle,
    listing options={language=Python,
                     basicstyle=\ttfamily\footnotesize,
                     xleftmargin=0mm,
                     breaklines=true, 
                     breakatwhitespace=true, 
                     commentstyle=\color{green!50!black}, 
                     showstringspaces=false,
                     },
    listing only,
    title=Code~\thetcbcounter,
    fonttitle=\footnotesize\sffamily,
    overlay unbroken and first={
        \path
            let
            \p1=(title.north east),
            \p2=(frame.north east)
            in
            node[anchor=west,font=\footnotesize\sffamily,color=bluPoli,text width=\x2-\x1]
            at (title.east) {#1};
    },
    label=#2
}

\begin{document}
\maketitle

\begin{abstract}
Hyperparameter optimization (HPO) is of paramount importance in the development of high-performance, specialized artificial intelligence (AI) models, ranging from well-established machine learning (ML) solutions to the deep learning (DL) domain and the field of spiking neural networks (SNNs).
The latter introduce further complexity due to the neuronal computational units and their additional hyperparameters, whose inadequate setting can dramatically impact the final model performance.
At the cost of possible reduced generalization capabilities, the most suitable strategy to fully disclose the power of SNNs is to adopt an application-oriented approach and perform extensive HPO experiments.
To facilitate these operations, automatic pipelines are fundamental, and their configuration is crucial.
In this document, the Neural Network Intelligence (NNI) toolkit is used as reference framework to present one such solution, with a use case example providing evidence of the corresponding results. In addition, a summary of published works employing the presented pipeline is reported as a potential source of insights into application-oriented HPO experiments for SNN prototyping.
\end{abstract}


\paragraph*{Declaration on generative artificial intelligence\\}
Revision for grammatical and spelling errors has been performed by means of \textit{gspeech}\footnote{\url{https://github.com/VitF/gspeech}}.

\section{Introduction}
It is nowadays well known to the general audience that \gls{ai} needs models to be trained. What may be less known to a general audience is what being trained actually implies in terms of the technical efforts required to be successful.
Any \gls{ai} model, whether it is in the \gls{ml} or \gls{dl} domain, can initially be broadly characterized by its trainable and non-trainable parameters.
In the case of neural networks, these can also be categorized into three main classes: weights, architecture configuration and training settings.
Despite being all of them inherently interconnected to each other, it is important to identify the different roles they have. By analogy, one could say that i) the architecture configuration \emph{builds} the model giving it the structure and the shape, ii) the training settings define how the model undergoes the learning stages and iii) the weights embed what the model learns through adaptation during the training procedure.
In a typical \gls{dl} case, only the weights are trained, although there may be exceptions based on specific user requirements.
What is not trained, typically referred to as the \emph{hyperparameters}, as opposed to the \emph{parameters} represented by the weights, then needs to be pre-defined and set, and this often represents a challenge towards high-performance models~\cite{liao_empirical_2022}.

In the neuromorphic computing domain, the use of \glspl{snn}, namely bio-inspired \glspl{ann} implementing brain-like primitives~\cite{Maass1997}, introduces further complexity due to the neuronal computational units and their additional hyperparameters~\cite{firmin_parallel_2024}.
\glspl{snn} are indeed built through simplified models of biological neurons, and even the simplest ones need for selection of at least one fundamental pair of hyperparameters: threshold voltage and decay constant~\cite{ou_overview_2022,ganguly_spike_2024}.
Inadequately chosen hyperparameters can result in suboptimal models with poor performance, which in turn can drive promising solutions to unsuccessful applications.
A possible way to avoid such outcomes is to perform \gls{hpo}~\cite{baratchi_automated_2024} with an application-oriented approach.
Although this can imply overspecialization taking shape in overfitting and poor generalization capability, a careful design of the \gls{hpo} experiment which takes into account the final target application, as well as its possible hardware-related constraints, is the most powerful solution when prototyping \glspl{snn}.

In this document, a pipeline for application-oriented automatic \gls{hpo} of \glspl{snn} is presented, relying on the \gls{nni}\footnote{github.com/microsoft/nni/} toolkit as reference framework.\\
The description is made such that the reader can go through a higher level discussion and a code-based example at the same time. The latter is sustained by a specific use case whose results are reported as well, and the proposed file names and structure retrace an actual \gls{hpo} experiment.

\section{Methodology}
The pipeline for automatic \gls{hpo} here proposed relies on the open-source \gls{nni} toolkit, which offers a variety of possibilities to carry out \gls{hpo} experiments as well as feature engineering, neural architecture search and model compression.
For application-oriented \gls{hpo}, it makes straightforward to define a trivial experiment capable of optimizing the objective metric of choice.
However, thanks to the richness of methods and tools it offers, it is also ideal for customization towards fine-tuned \gls{hpo} experiments capable of satisfying specific needs in terms of models, applications or metrics.

From a hierarchical perspective, the pipeline described in this document assumes that four constitutive \texttt{.py} files, the first being \texttt{nni\_main.py}, are at the same level within the root directory of the \gls{hpo} experiment.\\
As the first step in the direction of a fully user-defined experiment, the design of a properly customized configuration is crucial.
By means of \texttt{ExperimentConfig}, all the fundamental settings can be defined.
These include the name of the experiment and its directory, the command to run the main code containing the optimization objective, the search space to be explored for the identification of the optimal model, and how to carry out such exploration. Specifically, the latter must be defined in terms of a tuner, i.e. the optimization algorithm to be employed; then a duration of the experiment must be set together with early stopping conditions in case these are needed or wanted. Finally, the use of GPUs can be configured, and the number of trials to be run at the same time can be set if this is desired to be larger than 1.
All of these settings can be defined in the \texttt{nni\_main.py} file, where the \texttt{run} command is included too.

The \texttt{trial\_command} argument of \texttt{ExperimentConfig} in \texttt{nni\_main.py} is the driver to the next step, as it specifies what command to execute in order to make the experiment start.
As an example, the string \texttt{"python3 main\_hpo.py"} can be assigned to \texttt{trial\_command}, which means the \texttt{main\_hpo.py} script will be run for the \gls{hpo} experiment.
This second constitutive file can be thought of as the mediator between the \gls{nni} experiment infrastructure and the optimization objective, namely the model to be optimized. It indeed contains all the settings needed for the experiment except for the higher-level configurations related to how to address the optimization problem, which are instead defined in \texttt{nni\_main.py} only.
The ultimate goal of \texttt{main\_hpo.py} is the definition of what code to run, and with what settings, for the optimization problem. It embeds the fundamental information about the object to be optimized and the conditions for such optimization. Specifically, in the domain of \glspl{snn}, the \texttt{main\_hpo.py} file links the \gls{hpo} experiment configuration to the \gls{snn} model and the data to be used.
In the \texttt{main\_hpo.py} file, experiment management settings can also be included, such as a custom logging methodology and the saving criteria for the optimal models.
Alongside all these definitions, the most important command within \texttt{main\_hpo.py} is \texttt{nni.get\_next\_parameter()}, which ensures that the exploration of the search space defined in \texttt{nni\_main.py} is carried out by performing consecutive trials with different hyperparameter combinations.
The identification of the latter is managed by the \texttt{tuner} selected in \texttt{main\_hpo.py} through \texttt{ExperimentConfig}.

The aim of \texttt{main\_hpo.py} is hence to produce the objective metric to be optimized and to provide it to the \texttt{tuner}. This is pursued relying on two additional scripts: \texttt{setup\_hpo.py} and \texttt{train\_hpo.py}.
The first collects all the data-related aspects for the execution of the \gls{snn} model, as well as the specification of whether or not to use GPUs and the seed value; while the second defines the \gls{snn} itself and the training procedure.
All the specifics of the model being optimized, along with all custom operations during the learning phase, are in the \texttt{train\_hpo.py} file.
For instance, the handling of the validation step, if any, is defined here, as well as possible additional early stopping mechanisms to those provided by \gls{nni}.

With an overall view, the automatic \gls{hpo} pipeline can be described and summarized as \texttt{train\_hpo.py} being in charge of training a \gls{snn}-based model according to \texttt{setup\_hpo.py}, in order to provide \texttt{main\_hpo.py} with the information needed to solve the optimization problem described by \texttt{nni\_main.py}.

\subsection{Use case example}

In the following, by means of four code boxes corresponding to the files introduced above, a practical example of the pipeline is shown.
For the selected use case, a \gls{rsnn} is trained with the eligibility propagation (e-prop) algorithm~\cite{Bellec2020} partly relying on third-party code\footnote{github.com/ChFrenkel/eprop-PyTorch/}.
The task for this example is the Braille letter reading in its reduced form as adopted in~\cite{pedersen_neuromorphic_2024}.

The reported code is extracted from an actual \gls{hpo} project, but has been adapted to be embedded in this document. Consequently, criticalities may be present and direct usage without adjustments and proper validation is discouraged, as it might not result in the expected behaviour.
\newline

\begin{mycodebox}[HPO experiment configuration --- \texttt{nni\_main.py}]{code:HPO_conf}
from nni.experiment import *
import os
import sys

search_space = {
    'n_rec': {'_type': 'quniform', '_value': [11, 256, 1]},
    'threshold': {'_type': 'quniform', '_value': [0.05, 1, 0.05]},
    'tau_mem': {'_type': 'choice', '_value': [1e-3, 5e-3, 10e-3, 50e-3, 100e-3, 200e-3]},
    'tau_out': {'_type': 'choice', '_value': [1e-3, 5e-3, 10e-3, 50e-3, 100e-3, 200e-3]},
    'delay_targets': {'_type': 'choice', '_value': [1, 5, 10, 20, 50, 100]},
    'lr': {'_type': 'choice', '_value': [0.0001, 0.00015, 0.0002, 0.0005, 0.001, 0.0015, 0.002, 0.005, 0.01, 0.02, 0.05, 0.1]},
    'gamma': {'_type': 'quniform', '_value': [0.1, 1, 0.1]},
    'reset_mechanism': {'_type': 'choice', '_value': ["subtract", "zero"]},
}

exp_name = "ReckOn_braille_nir"
searchspace_path = os.path.join("./",exp_name)
with open(searchspace_path, "w") as write_searchspace:
    json.dump(search_space, write_searchspace)

config = ExperimentConfig(
    experiment_name = exp_name,
    experiment_working_directory = f"~/nni-experiments/{exp_name}",
    trial_command = "python3 main_hpo.py",
    trial_code_directory = "./",
    search_space = search_space,
    tuner = AlgorithmConfig(name="Anneal",
                            class_args={"optimize_mode": "maximize"}),
    assessor = AlgorithmConfig(name="Medianstop",
                               class_args=({'optimize_mode': 'maximize',
                                            'start_step': 10})),
    tuner_gpu_indices = [0,1],
    max_trial_number = 1000,
    max_experiment_duration = "100d",
    trial_concurrency = 2,
    training_service = LocalConfig(trial_gpu_number=2,
                                   max_trial_number_per_gpu=3,
                                   use_active_gpu=True)
)

experiment = Experiment(config)

experiment.run(8080)

# Stop through input
input('Press any key to stop the experiment.')

# Stop at the end
experiment.stop()
\end{mycodebox}

\begin{mycodebox}[HPO experiment startup and coordination --- \texttt{main\_hpo.py} partly based on \href{github.com/ChFrenkel/eprop-PyTorch/main.py}{third-party code}]{code:HPO_startup}
import argparse
from collections import namedtuple
from copy import deepcopy
import datetime
import logging
import nni
from nni.tools.nnictl import updater
import numpy as np
import os
import pickle as pkl
import sys
import torch

from nni_main import args as args_main
from nni_main import searchspace_path
import train_hpo
import setup_hpo

class SearchSpaceUpdater(object):
    def __init__(self, *initial_data, **kwargs):
        for dictionary in initial_data:
            for key in dictionary:
                setattr(self, key, dictionary[key])
        for key in kwargs:
            setattr(self, key, kwargs[key])


def main():

    parser = argparse.ArgumentParser(description='Spiking RNN PyTorch training')
    ## General
    parser.add_argument('--nni_hpo', type=bool, default=True, help='Specify if HPO is being run or not')
    parser.add_argument('--cpu', action='store_true', default=False, help='Disable CUDA training and run training on CPU')
    parser.add_argument('--manual_gpu_idx', type=int, default=[0,1], help='Set which GPU to use.')
    parser.add_argument('--gpu_mem_frac', type=float, default=0.95, help='The maximum GPU memory fraction to be used by this experiment.')
    parser.add_argument('--dataset', type=str, choices=["braille_nir"], default="braille_nir", help='Choice of the dataset')
    parser.add_argument('--shuffle', type=bool, default=True, help='Enables shuffling sample order in datasets after each epoch')
    parser.add_argument('--trials', type=int, default=1, help='Number of trial experiments to do (i.e. repetitions with different initializations)')
    parser.add_argument('--epochs', type=int, default=1000, help='Number of epochs to train')
    parser.add_argument('--optimizer', type=str, choices = ['Adam'], default='Adam', help='Choice of the optimizer')
    parser.add_argument('--loss', type=str, choices = ['CE'], default='CE', help='Choice of the loss function')
    parser.add_argument('--lr', type=float, default=1e-3, help='Initial learning rate')
    parser.add_argument('--lr-layer-norm', type=float, nargs='+', default=(0.05,0.05,1.0), help='Per-layer modulation factor of the learning rate')
    parser.add_argument('--batch-size', type=int, default=10, help='Batch size for training.')
    parser.add_argument('--val-batch-size', type=int, default=10, help='Batch size for validation.')
    parser.add_argument('--test-batch-size', type=int, default=10, help='Batch size for test.')
    parser.add_argument('--train-len', type=int, default=8189, help='Number of training set samples')
    parser.add_argument('--val-len', type=int, default=2994, help='Number of validation set samples')
    parser.add_argument('--test-len', type=int, default=3116, help='Number of test set samples')
    parser.add_argument('--visualize', type=bool, default=False, help='Enable network visualization')
    parser.add_argument('--visualize-light', type=bool, default=False, help='Enable light mode in network visualization, plots traces only for a single neuron')
    parser.add_argument('--log_dir', type=str, default="./logs")
    parser.add_argument('--model_dir', type=str, default="./models")
    parser.add_argument('--report_dir', type=str, default="./reports")
    parser.add_argument('--result_dir', type=str, default="./results")
    parser.add_argument('--save_model', type=bool, default=True)
    parser.add_argument('--store_model', type=bool, default=True)
    ## Network model parameters
    parser.add_argument('--n_rec', type=int, default=64, help='Number of recurrent units')
    parser.add_argument('--model', type=str, choices = ['LIF'], default='LIF', help='Neuron model in the recurrent layer.')
    parser.add_argument('--threshold', type=float, default=0.9, help='Firing threshold in the recurrent layer')
    parser.add_argument('--tau-mem', type=float, default=250e-3, help='Membrane potential leakage time constant in the recurrent layer (in seconds)')
    parser.add_argument('--tau-out', type=float, default=5e-3, help='Membrane potential leakage time constant in the output layer (in seconds)')
    parser.add_argument('--bias-out', type=float, default=0.0, help='Bias of the output layer')
    parser.add_argument('--gamma', type=float, default=0.3, help='Surrogate derivative magnitude parameter')
    parser.add_argument('--w-init-gain', type=float, nargs='+', default=(0.5,0.1,0.5), help='Gain parameter for the He Normal initialization of the input, recurrent and output layer weights')
    
    args = parser.parse_args()

    args.experiment_name = args_main.exp_name
    args.searchspace_path = searchspace_path
    args.visible_gpus = args_main.exp_gpu_sel

    (device, train_loader, val_loader, test_loader, LOG) = setup_hpo.setup(args)

    try:

        trial_datetime = datetime.datetime.now().strftime("%Y%m%d_%H%M%S")

        LOG.debug("----------------------------------------")
        LOG.debug("\n")
        LOG.debug("Trial {} (# {}, ID {}) started on: {}-{}-{} {}:{}:{}\n".format(
            nni.get_sequence_id()+1,
            nni.get_sequence_id(),
            nni.get_trial_id(),
            trial_datetime[:4],
            trial_datetime[4:6],
            trial_datetime[6:8],
            trial_datetime[-6:-4],
            trial_datetime[-4:-2],
            trial_datetime[-2:]))
    
        ### Every n_tr trials, "update" the searchspace inducing a new RandomState for the tuner
        n_tr = 250
        update_searchspace = SearchSpaceUpdater({"filename": searchspace_path, "id": nni.get_experiment_id()})
        if (nni.get_sequence_id() > 0) & (nni.get_sequence_id()\%n_tr == 0):
            updater.update_searchspace(update_searchspace) # it will use update_searchspace.filename to update the search space
        
        settings = vars(args)
        ### Get parameters from the tuner combining them with the line arguments
        settings_nni = nni.get_next_parameter()
        for ii in settings_nni.keys():
            if ii in settings.keys():
                del settings[ii]
    
        PARAMS = {**settings, **settings_nni}

        LOG.debug("Parameters selected for trial {} (# {}, ID {}): {}\n".format(
            nni.get_sequence_id()+1, nni.get_sequence_id(), nni.get_trial_id(), PARAMS))

        DictObj = namedtuple('DictObject', PARAMS.keys())
        PARAMS = DictObj(**PARAMS)
    
        LOG.debug("\n")

        overall_results, test_best_val, best_val_model = train_hpo.train(PARAMS, device, train_loader, val_loader, test_loader, LOG)
        
        ### Report results (i.e. test accuracy from best validation) of each trial
        report_path = os.path.join(args.report_dir,
            args.experiment_name,
            f"{nni.get_experiment_id()}")
        with open(os.path.join(report_path,"report_test"), 'a') as f:
            f.write("{} % \t test accuracy ({})".format(test_best_val,nni.get_trial_id()))
            f.write('\n')
        
        ### Save trained weights giving the highest test accuracy
        if args.save_model:
            save_model_path = os.path.join(args.model_dir,
                args.experiment_name,
                f"{nni.get_experiment_id()}")
            with open(os.path.join(report_path,"report_test"), 'r') as f:
                if test_best_val >= np.max(np.asarray([(line.strip().split(" ")[0]) for line in f], dtype=np.float64)):
                    torch.save(deepcopy(best_val_model),
                        os.path.join(save_model_path,
                            f"{nni.get_sequence_id()}\
                            _{trial_datetime}\
                            _{nni.get_trial_id()}"))
                    # And save the overall results collected for such model
                    results_path = os.path.join(args.result_dir,
                        args.experiment_name,
                        nni.get_experiment_id())
                    with open(os.path.join(results_path,
                        f"{nni.get_trial_id()}_train_acc.pkl"),
                        'wb') as f:
                        pkl.dump(overall_results[1], f)
                    with open(os.path.join(results_path,
                        f"{nni.get_trial_id()}_train_loss.pkl"),
                        'wb') as f:
                        pkl.dump(overall_results[0], f)
                    if not args.skip_val:
                        with open(os.path.join(results_path,
                            f"{nni.get_trial_id()}_val_acc.pkl"),
                            'wb') as f:
                            pkl.dump(overall_results[3], f)
                        with open(os.path.join(results_path,
                            f"{nni.get_trial_id()}_val_loss.pkl"),
                            'wb') as f:
                            pkl.dump(overall_results[2], f)
                    if not args.skip_test:
                        with open(os.path.join(results_path,
                            f"{nni.get_trial_id()}_test_acc.pkl"),
                            'wb') as f:
                            pkl.dump(overall_results[5], f)
                        with open(os.path.join(results_path,
                            f"{nni.get_trial_id()}_test_loss.pkl"),
                            'wb') as f:
                            pkl.dump(overall_results[4], f)
    
    except Exception as e:
        LOG.exception(e)
        raise

    LOG.debug("=== Trial completed ===")
    print(f"=== TRIAL #{nni.get_sequence_id()} DONE ===")

if __name__ == '__main__':
    main()
\end{mycodebox}

\begin{mycodebox}[HPO setup --- \texttt{setup\_hpo.py} partly based on \href{github.com/ChFrenkel/eprop-PyTorch/setup\_hpo.py}{third-party code}]{code:HPO_setup}
import datetime
import logging
import numpy as np
import numpy.random as rd
import os
import sys
import torch

def setup(args):

    experiment_datetime = datetime.datetime.now().strftime("%Y%m%d_%H%M%S")
    args.experiment_datetime = experiment_datetime

    ### log file configuration ###########################################

    log_path = os.path.join(args.log_dir,args.experiment_name)
    create_directory(log_path)
    LOG = logging.getLogger(args.experiment_name)
    logging.basicConfig(filename=log_path+"/{}.log".format(experiment_datetime),
                        filemode='a',
                        format="%(asctime)s %(name)s %(message)s",
                        datefmt='%Y%m%d_%H%M%S')
    LOG.setLevel(logging.DEBUG)
    LOG.debug(f"{args.experiment_name} HPO experiment\n")
    LOG.debug("Experiment started on: {}-{}-{} {}:{}:{}\n".format(
        experiment_datetime[:4],
        experiment_datetime[4:6],
        experiment_datetime[6:8],
        experiment_datetime[-6:-4],
        experiment_datetime[-4:-2],
        experiment_datetime[-2:])
        )
    
    ######################################################################

    args.cuda = not args.cpu and torch.cuda.is_available()
    if args.cuda:
        device = torch.device(f'cuda:{args.manual_gpu_idx}')     
    else:
        device = torch.device('cpu')
    
    kwargs = {'num_workers': 0, 'pin_memory': True} if args.cuda else {}

    if args.dataset == "braille_nir":

        print("=== Loading the Braille dataset from NIR...")
        if args.nni_hpo:
            random_split = np.random.randint(0,10)
            LOG.debug(f"Training-validation split used: {random_split}")
        else:
            random_split = None

        (train_loader, val_loader, test_loader, sizes) = braille_data_nir(dataDir=f"../data/{args.dataset}", device=device, batch_size=args.batch_size, shuffle=True, random_split=random_split)
        print("=== ...loading completed:")

        args.n_classes                  = sizes[-1]
        args.n_steps, args.n_inputs     = tuple(next(iter(train_loader))[0].shape[1:])
        args.dt                         = 1e-3
        args.classif                    = True
        args.full_train_len             = sizes[0]
        args.full_val_len               = sizes[1]
        args.full_test_len              = sizes[2]
        args.skip_val                   = False
        args.skip_test                  = False

        print("\ttraining set length: " + str(args.full_train_len))
        print("\tvalidation set length: " + str(args.full_val_len))
        print("\ttest set length: " + str(args.full_test_len))

        seed = 42
        if args.cuda:
            os.environ['PYTHONHASHSEED'] = str(seed)
            os.environ['CUBLAS_WORKSPACE_CONFIG'] = ":4096:8"
        np.random.seed(seed)
        random.seed(seed)
        torch.manual_seed(seed)
        if seed != None:
            LOG.debug("\nSeed set to {}\n".format(seed))

        return (device, train_loader, val_loader, test_loader, LOG)
\end{mycodebox}

\begin{mycodebox}[HPO objective --- \texttt{train\_hpo.py} partly based on \href{github.com/ChFrenkel/eprop-PyTorch/train\_hpo.py}{third-party code}]{code:HPO_train}
from copy import deepcopy
import nni
import numpy as np
import os
import pickle as pkl
import sys
import torch
import torch.nn as nn
import torch.nn.functional as F
import torch.optim as optim

import models_hpo # CUSTOM MODELS


def train(args, device, train_loader, traintest_loader, test_loader, LOG):

    models_path = os.path.join(args.model_dir,args.experiment_name,"tmp")
    create_directory(models_path)

    train_acc_rec = []
    train_loss_rec = []
    val_acc_rec = []
    val_loss_rec = []
    test_acc_rec = []
    test_loss_rec = []
    
    for trial in range(1,args.trials+1):

        if args.dataset == "braille_nir":
            if args.reset_mechanism == "zero":
                n_rec = 38
            elif args.reset_mechanismreset_mechanism == "subtract":
                n_rec = 40
        
        # Network topology
        model = models_hpo.SRNN(n_in=args.n_inputs,
                            n_rec=n_rec,
                            n_out=args.n_classes,
                            n_t=args.n_steps,
                            thr=args.threshold,
                            tau_m=args.tau_mem,
                            tau_o=args.tau_out,
                            b_o=args.bias_out,
                            gamma=args.gamma,
                            reset_mechanism=args.reset_mechanism,
                            dt=args.dt,
                            model=args.model,
                            classif=args.classif,
                            w_init_gain=args.w_init_gain,
                            lr_layer=args.lr_layer_norm,
                            t_crop=args.delay_targets,
                            visualize=args.visualize,
                            visualize_light=args.visualize_light,
                            device=device)

        # Use CUDA for GPU-based computation if enabled
        if args.cuda:
            model.cuda(device)
        
        # Initial monitoring
        if (args.trials > 1):
            LOG.debug('\nIn trial {} of {}'.format(trial,args.trials))
        if (trial == 1):
            LOG.debug("=== Model ===")
            LOG.debug(f"{model}\n")
        
        # Optimizer
        if args.optimizer == 'Adam':
            optimizer = optim.Adam(model.parameters(), lr=args.lr)
        
        # Loss function (only for performance monitoring purposes, does not influence learning as e-prop learning is hardcoded)
        if args.loss == 'CE':
            loss = (F.cross_entropy, (lambda l : torch.max(l, 1)[1]))
        
        ### Training and validation with a while loop including early stopping
        LOG.debug("=== Starting model training with %d epochs:\n" % (args.epochs,))
        epoch = 0
        EarlyStop_delta_val_loss = 0.5 # intended as the percentage of change in validation loss
        counter_small_delta_loss = 0
        stop_small_delta_loss = 10 # how many times the condition must be met to induce early stopping
        EarlyStop_delta_val_loss_up = 0.5 # intended as the percentage of increase in validation loss
        counter_delta_loss_up = 0
        stop_delta_loss_up = 10 # how many times the condition must be met to induce early stopping
        EarlyStop_delta_val_acc_low = 0.1 # intended as the percentage of change in validation accuracy
        counter_small_delta_acc = 0
        stop_small_delta_acc = 5 # how many times the condition must be met to induce early stopping
        EarlyStop_delta_val_acc_high = 2 # intended as the percentage of decrease in validation accuracy
        counter_large_delta_acc = 0
        stop_large_delta_acc = 5 # how many times the condition must be met to induce early stopping

        while (counter_small_delta_loss < stop_small_delta_loss) & (counter_delta_loss_up < stop_delta_loss_up) & (counter_small_delta_acc < stop_small_delta_acc) & (counter_large_delta_acc < stop_large_delta_acc) & (epoch < args.epochs):
        
            epoch += 1
            LOG.debug(f"\t Epoch {epoch}/{args.epochs}...")

            if args.classif:

                #Training:
                train_acc, train_loss = do_epoch(args, True, model, device, train_loader, optimizer, loss, 'train', LOG)
                train_acc_rec.append(train_acc)
                train_loss_rec.append(train_loss)
                
                # Check performance on the training set and on the validation set:
                if not args.skip_val:
                    val_acc, val_loss = do_epoch(args, False, model, device, traintest_loader, optimizer, loss, 'val', LOG)
                    val_acc_rec.append(val_acc)
                    val_loss_rec.append(val_loss)
                    if val_acc >= max(val_acc_rec):
                        best_val_acc = val_acc
                        best_val_epoch = epoch
                        torch.save(deepcopy(model), 
                            os.path.join(models_path,
                                f"{args.experiment_datetime}_best_val"))
                    nni.report_intermediate_result({
                        "default": round(val_acc,4),
                        "training acc.": round(train_acc,4),
                        "val. loss": round(val_loss,5),
                        "train. loss": round(train_loss,5)})
                
                if epoch >= 2:
                    # Check loss variations (on validation data) during training: count number of epoch with SMALL (< EarlyStop_delta_val_loss %) CHANGES
                    if abs(val_loss_rec[-1] - val_loss_rec[-2])/val_loss_rec[-2]*100 < EarlyStop_delta_val_loss:
                        counter_small_delta_loss += 1
                    else:
                        counter_small_delta_loss = 0
                    # Check loss variations (on validation data) during training: count number of epoch with LARGE (> EarlyStop_delta_val_loss_up %) INCREASE
                    if (val_loss_rec[-1] - val_loss_rec[-2])/val_loss_rec[-2]*100 > EarlyStop_delta_val_loss_up:
                        counter_delta_loss_up += 1
                    else:
                        counter_delta_loss_up = 0
                    # check accuracy variations (on validation data) during training: count number of epoch with SMALL (> EarlyStop_delta_val_acc %) CHANGES
                    if abs(val_acc_rec[-1] - val_acc_rec[-2])/val_acc_rec[-2]*100 < EarlyStop_delta_val_acc_low:
                        counter_small_delta_acc += 1
                    else:
                        counter_small_delta_acc = 0
                    # check accuracy variations (on validation data) during training: count number of epoch with LARGE (> EarlyStop_delta_val_acc %) DECREASE
                    if (val_acc_rec[-2] - val_acc_rec[-1])/val_acc_rec[-2]*100 > EarlyStop_delta_val_acc_high:
                        counter_large_delta_acc += 1
                    else:
                        counter_large_delta_acc = 0
        
        if not args.skip_test:
            LOG.debug(f"--- loading model with best validation accuracy ({round(best_val_acc,4)}% at epoch {best_val_epoch}) ---")
            model_test = torch.load(os.path.join(models_path,
                f"{args.experiment_datetime}_best_val"))
            test_acc, test_loss = do_epoch(args, False, model_test, device, test_loader, optimizer, loss, 'test', LOG)
            test_acc_rec.append(test_acc)
            test_loss_rec.append(test_loss)
        
        LOG.debug("\n")
        if counter_small_delta_loss >= stop_small_delta_loss:
            LOG.debug("Training stopped after {}/{} epochs: stop condition for small validation loss changes met.".format(epoch,args.epochs))
        elif counter_delta_loss_up >= stop_delta_loss_up:
            LOG.debug("Training stopped after {}/{} epochs: stop condition for validation loss increase met.".format(epoch,args.epochs))
        elif counter_small_delta_acc >= stop_small_delta_acc:
            LOG.debug("Training stopped after {}/{} epochs: stop condition for small validation accuracy changes met.".format(epoch,args.epochs))
        elif counter_large_delta_acc >= stop_large_delta_acc:
            LOG.debug("Training stopped after {}/{} epochs: stop condition for validation accuracy decrease met.".format(epoch,args.epochs))
        else:
            LOG.debug("Training ended after {}/{} epochs.".format(epoch,args.epochs))
        
        # best training and validation at best training
        acc_best_train = np.max(train_acc_rec)
        epoch_best_train = np.argmax(train_acc_rec)
        acc_val_at_best_train = val_acc_rec[epoch_best_train]

        # best validation and training at best validation
        acc_best_val = np.max(val_acc_rec)
        epoch_best_val = np.argmax(val_acc_rec)
        acc_train_at_best_val = train_acc_rec[epoch_best_val]

        LOG.debug("\n")
        LOG.debug("Trial results: ")
        LOG.debug("\tBest training accuracy: {}% ({}% corresponding validation accuracy) at epoch {}/{}".format(
            np.round(acc_best_train,4), np.round(acc_val_at_best_train,4), epoch_best_train+1, args.epochs))
        LOG.debug("\tBest validation accuracy: {}% ({}% corresponding training accuracy) at epoch {}/{}".format(
            np.round(acc_best_val,4), np.round(acc_train_at_best_val,4), epoch_best_val+1, args.epochs))
        LOG.debug("\tTest accuracy (from best validation): {}%".format(
            np.round(test_acc,4)))
        LOG.debug("\n")

        LOG.debug("----------------------------------------\n\n")

        nni.report_final_result({"default": np.round(acc_best_val,4), # the default value is the maximum validation accuracy achieved
                                 "best training": np.round(acc_best_train,4),
                                 "test": np.round(test_acc,4)})
    
    return [train_loss_rec, train_acc_rec, val_loss_rec, val_acc_rec, test_loss_rec, test_acc_rec], test_acc, model_test


def do_epoch(args, do_training, model, device, loader, optimizer, loss_fct, benchType, LOG):

    model.eval()    # This implementation does not rely on autograd, learning update rules are hardcoded
    
    score = 0
    loss = 0

    if benchType == "train":
        batch = args.batch_size
        length = args.full_train_len
    elif benchType == "val":
        batch = args.val_batch_size
        length = args.full_val_len
    elif benchType == "test":
        batch = args.test_batch_size
        length = args.full_test_len

    if benchType == "val":
        LOG.debug("\t --> Validation")
    elif benchType == "test":
        LOG.debug("\t ### Test ###")
    
    with torch.no_grad():   # Same here, we make sure autograd is disabled
        
        # For each batch
        for batch_idx, (data, label) in enumerate(loader):

            if args.classif:    # Do a one-hot encoding for classification
                targets = torch.zeros(label.shape, 
                    device=device)[:,None,None].expand(-1,-1,
                        args.n_classes).scatter(2,label[:,None,None], 1.0).permute(1,0,2)
                targets = torch.tile(targets,(data.shape[1],1,1))
            else:
                targets = label.permute(1,0,2)

            # Evaluate the model for all the time steps of the input data, then either do the weight updates on a per-timestep basis, or on a per-sample basis (sum of all per-timestep updates).
            optimizer.zero_grad()
            output = model(data.permute(1,0,2), targets, do_training)
            if do_training:
                optimizer.step()
                
            # Compute the loss function, inference and score
            if args.delay_targets:
                loss += loss_fct[0](output.detach()[-args.delay_targets:], loss_fct[1](targets[-args.delay_targets:]), reduction='mean')
            else:
                loss += loss_fct[0](output.detach(), loss_fct[1](targets), reduction='mean')
            if args.classif:
                if args.delay_targets:
                    inference = torch.argmax(torch.sum(output.detach()\
                        [-args.delay_targets:],axis=0),axis=1)
                    score += torch.sum(torch.eq(inference,label))
                else:
                    inference = torch.argmax(torch.sum(output.detach(),axis=0),axis=1)
                    score += torch.sum(torch.eq(inference,label))
        
    if benchType == "train" and do_training:
        info = "on training set (while training): "
    elif benchType == "train":
        info = "on training set                 : "
    elif benchType == "val":
        info = "on validation set               : "
    elif benchType == "test":
        info = "on test set                     : "

    if args.classif:
        LOG.debug("\t\t Score " + info + str(score.item()) + '/' + str(length) + ' (' + str(round(score.item()/length*100,4)) + '%), loss: ' + str(round(loss.item(),4)))
        return score.item()/length*100, loss.item()
    else:
        LOG.debug("\t\t Loss " + info + str(round(loss.item(),4)))
        return loss.item()
\end{mycodebox}

\section{Results}

\subsection{Use case example}

\begin{figure}[b]
    \centering
    \includegraphics[width=\textwidth]{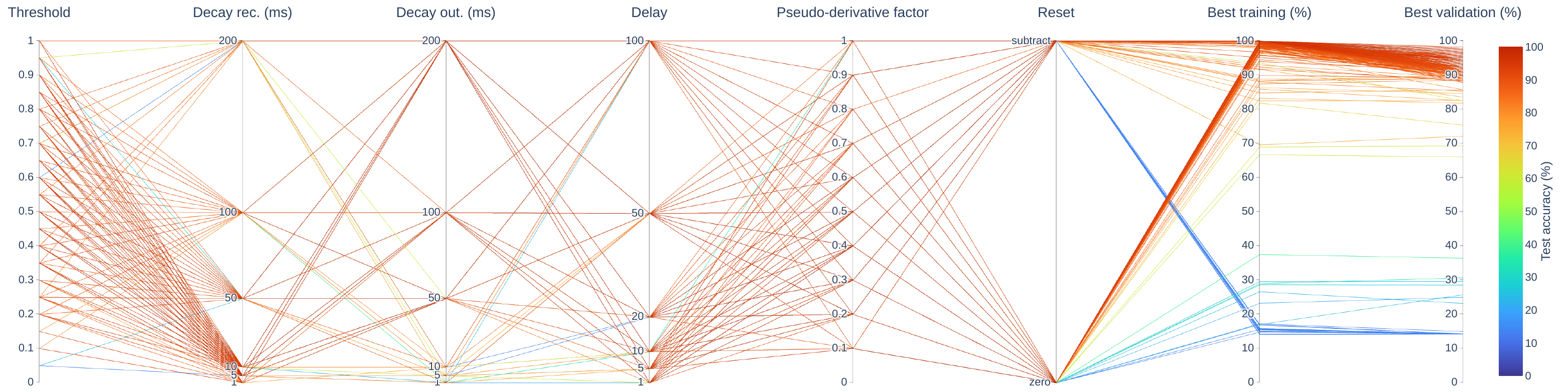}
    \caption{Exploration of the search space with the resulting test accuracy for each combination of hyperparameters}
    \label{fig:par_coord}
\end{figure}

\begin{figure}[t]
    \centering
    \includegraphics[width=0.5\textwidth]{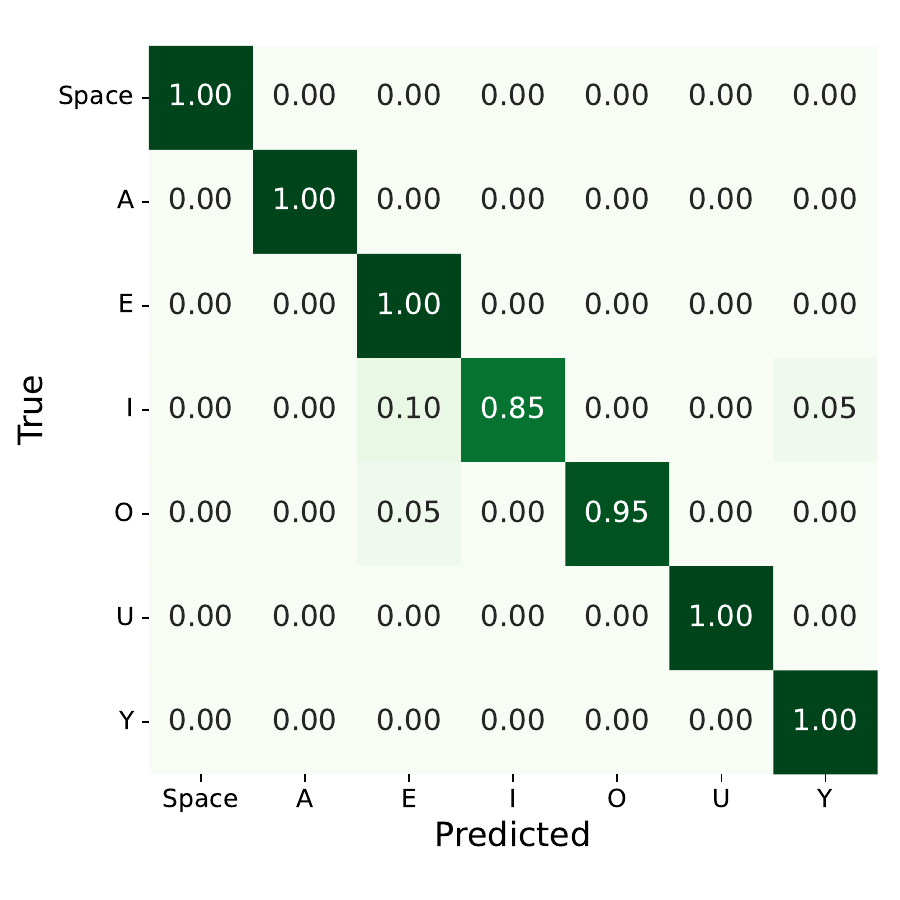}
    \caption{Confusion matrix produced on the test set by the optimal model. The overall accuracy is 97.14\%.}
    \label{fig:cm}
\end{figure}

The \gls{hpo} experiment run using the scripts referred to in the code boxes above results, at the time of writing, in the state-of-the-art classification of Braille characters from the reduced dataset used in~\cite{pedersen_neuromorphic_2024}. Figure~\ref{fig:par_coord} summarizes the exploration carried out across the search space defined in Code~\ref{code:HPO_conf}, reporting both the best training accuracy and the best validation accuracy achieved during the learning phase of each trial.
In the rightmost part, the test accuracy is reported.
As it is reported in Code~\ref{code:HPO_train}, the value for the \texttt{default} key in the dictionary given to \texttt{nni.report\_final\_result()}, namely the objective metric for optimization with this experiment, is set to be the validation accuracy; particularly, the best validation accuracy achieved throughout the training epochs.\\
From Code~\ref{code:HPO_train}, it is also possible to track down the selection criterion for the optimal model.
At the end of the training stage, the weights from the highest validation accuracy are loaded, and test is performed.
The resulting accuracy is passed to \texttt{nni.report\_final\_result()} as value for the key \texttt{test}, and the best test accuracy at the end of the \gls{hpo} experiment will identify what combination of hyperparameters is the optimal one for the model under optimization in the selected task.\\
In Figure~\ref{fig:cm}, the confusion matrix produced on the test set by the optimized \gls{snn} is shown, with partial misclassification in two classes only and an overall accuracy of 97.14\%.

\subsection{Published works}

The application-oriented automatic \gls{hpo} procedure described in this document is the result of ongoing efforts that lead to continuous refinement and customization of the pipeline initially proposed in~\cite{fra_human_2022}.
Its adaptability, rooted in the wide range of possibilities offered by \gls{nni}, is at the same time the key feature for its employment and the driving force for its never-ending development. In Table~\ref{table:works}, a summary of the published works that use it for spiking models is reported.

\begin{table*}[h]
    \renewcommand{\arraystretch}{1.15}
    \centering
    \caption{Summary of published works that performed application-oriented automatic \gls{hpo} through \gls{nni} based on the procedure presented here}
    \label{table:works}
    \begin{tabular}{{|>{\centering}m{2cm}|>{\centering}m{2.7cm}|>{\centering}m{2.5cm}|>{\centering}m{1.9cm}|>{\centering}m{2cm}|>{\centering\arraybackslash}m{1.8cm}|}}
        \hline
        { \textbf{Reference} } & { \textbf{Task} } & { \textbf{Architecture} } & { \textbf{Event/Frame data} } & { \textbf{Dataset} } & { \textbf{Framework} } \\
        \hline
        { \cite{fra_human_2022} } & { Human activity recognition } & { LMU } & { Frame } & { \cite{Weiss2019a,Weiss2019} } & { \texttt{TensorFlow} } \\
        \hline
        { \cite{muller-cleve_braille_2022} } & { Braille letter reading } & { Fully connected } & { Both } & { \cite{muller-cleve_tactile_2022} } & { \texttt{PyTorch} } \\
        \hline
        { \cite{pedersen_neuromorphic_2024} } & { Braille letter reading } & { Fully connected } & { Event } & { \href{https://github.com/neuromorphs/NIR/tree/main/paper/03_rnn/data}{Braille subset for \cite{pedersen_neuromorphic_2024}} } & { \texttt{snnTorch} } \\
        \hline
        { \cite{wand_natively_2024} } & { Human activity recognition } & { L$^2$MU } & { Frame } & { \cite{Weiss2019a,Weiss2019} } & { \texttt{snnTorch} } \\
        \hline
        { \cite{meo_neu-brauer_2025} } & { Braille letter reading } & { Fully connected } & { Frame } & { \cite{muller-cleve_tactile_2022} } & { \texttt{snnTorch} } \\
        \hline
        { \cite{fra_win-gui_2025} } & { Spike pattern classification } & { Fully connected } & { Event } & { Spike patterns from \cite{Mihalas2009} } & { \texttt{snnTorch} } \\
        \hline
        { \cite{leto_lif-based_2025} } & { Braille letter reading } & { L$^2$MU } & { Event } & { \cite{muller-cleve_tactile_2022} and \href{https://github.com/neuromorphs/NIR/tree/main/paper/03\_rnn/data}{Braille subset} for \cite{pedersen_neuromorphic_2024} } & { \texttt{snnTorch} } \\
        \hline
        { \cite{leto_variable-precision_2026} } & { Human activity recognition } & { L$^2$MU } & { Frame } & { \href{https://huggingface.co/datasets/neuromorphic-polito/siddha}{SIDDHA} } & { \texttt{snnTorch} } \\
        \hline
    \end{tabular}
\end{table*}

\section*{Acknowledgments}
This Research is funded by the European Union - NextGenerationEU Project 3A-ITALY MICS (PE0000004, CUP E13C22001900001, Spoke 6).

\bibliographystyle{unsrt}  
\bibliography{HPO_arXiv.bib}

\end{document}